\documentclass[sigconf]{acmart}
\AtBeginDocument{%
  }

\setcopyright{acmlicensed}
\copyrightyear{2026}
\acmYear{2026}
\acmDOI{XXXXXXX.XXXXXXX}
\acmConference[3rd HEAL Workshop - CHI]{April,
  2026}{Barcelona, Spain}

\begin{document}


\title[LLM-Generated Lessons Evaluation]{Evaluating LLM-Generated Lessons from the Language Learning Students' Perspective: A Short Case Study on Duolingo}


\author{Carlos Rafael Catalan}
\affiliation{%
  \institution{Samsung R\&D Institute Philippines}
  \city{Manila}
  \country{Philippines}
  }
  \email{c.catalan@samsung.com}

  \author{Patricia Nicole Monderin}
\affiliation{%
  \institution{Samsung R\&D Institute Philippines}
  \city{Manila}
  \country{Philippines}
  }
  \email{p.monderin@samsung.com}

\author{Lheane Marie Dizon}
\affiliation{%
  \institution{Samsung R\&D Institute Philippines}
  \city{Manila}
  \country{Philippines}
  }
  \email{lm.dizon@samsung.com}

\author{Gap Estrella}
\affiliation{%
  \institution{Samsung R\&D Institute Philippines}
  \city{Manila}
  \country{Philippines}
  }
  \email{pg.estrella@samsung.com}

\author{Raymund John Sarmiento}
\affiliation{%
  \institution{Samsung R\&D Institute Philippines}
  \city{Manila}
  \country{Philippines}
  }
  \email{rj.sarmiento@samsung.com}

\author{Marie Antoinette Patalagsa}
\affiliation{%
  \institution{Samsung R\&D Institute Philippines}
  \city{Manila}
  \country{Philippines}
  }
\email{m.patalagsa@samsung.com}

\renewcommand{\shortauthors}{Catalan et al.}

\begin{abstract}
   Popular language learning applications such as Duolingo use large language models (LLMs) to generate lessons for its users. Most lessons focus on general real-world scenarios such as greetings, ordering food, or asking directions, with limited support for profession-specific contexts. This gap can hinder learners from achieving professional-level fluency, which we define as the ability to communicate comfortably various work-related and domain-specific information in the target language. We surveyed five employees from a multinational company in the Philippines on their experiences with Duolingo. Results show that respondents encountered general scenarios more frequently than work-related ones, and that the former are relatable and effective in building foundational grammar, vocabulary, and cultural knowledge. The latter helps bridge the gap toward professional fluency as it contains domain-specific vocabulary. Each participant suggested lesson scenarios that diverge in contexts when analyzed in aggregate. With this understanding, we propose that language learning applications should generate lessons that adapt to an individual’s needs through personalized, domain-specific lesson scenarios while maintaining foundational support through general, relatable lesson scenarios.
\end{abstract}





\maketitle

\section{Introduction}

Large Language Models (LLMs) show exceptional capabilities in educational use-cases, such as generating lessons for students \cite{mogavi2024chatgpt, mollick2023using, zheng2025knowledge}. In the domain of language learning, this technology has made language acquisition applications such as Duolingo \cite{duolingo-ai} a popular tool for users to acquire fluency in another language \cite{duolingo-philippines, duolingo-report}. In line with traditional language learning settings, generated lessons typically depict real-world scenarios in the target language to immerse the student to help them gain proficiency and fluency\cite{krashen1981second}. 


\begin{figure}[h]
  \centering
  \includegraphics[width=\linewidth]{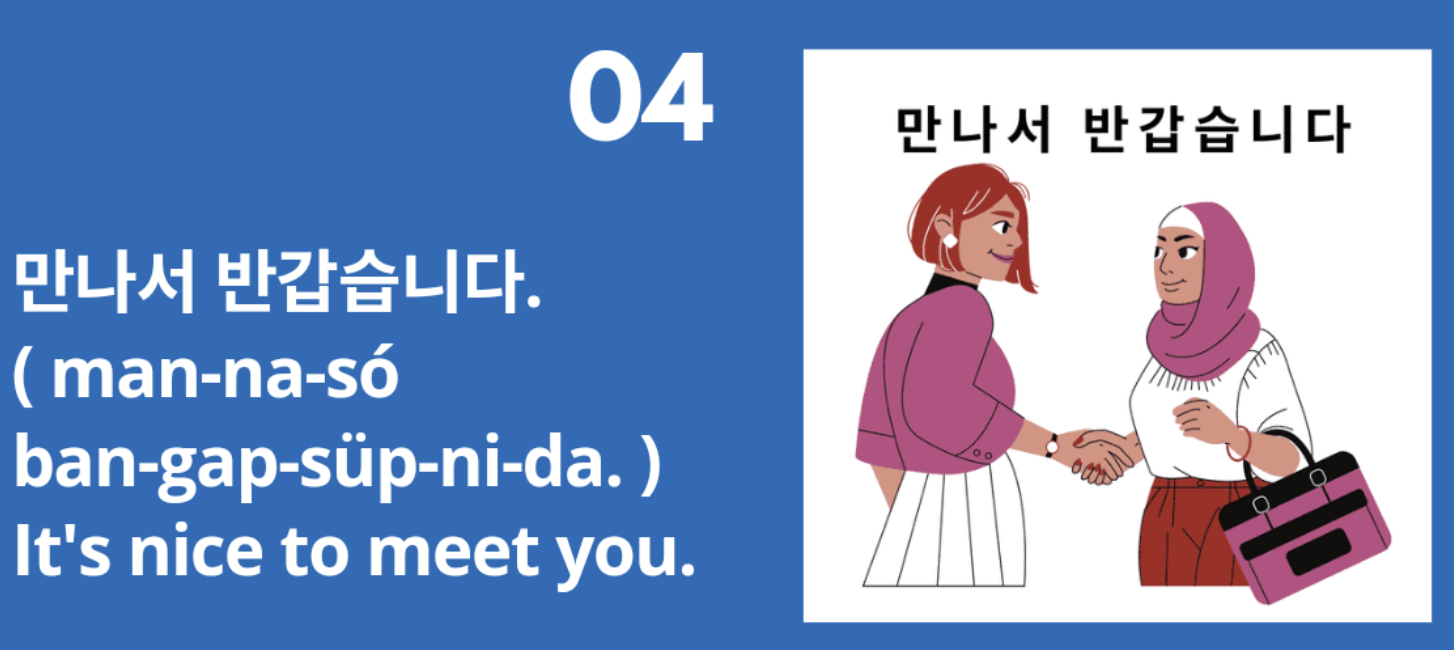}
  \caption{A scenario showing a greeting in Korean. Presenting a real-world scenario is typical of language learning resources, as it is able to immerse the learner in the target language. (Image was grabbed from https://www.topikguide.com/korean-greetings/)}
  \Description{A scenario showing a greeting in Korean. Presenting a real-world scenario is typical of language learning resources, as it is able to immerse the learner in the target language.}
\end{figure}

To accommodate students of diverse backgrounds, these lessons often depict general scenarios, such as greetings, ordering food, and asking for directions. However, very rarely do these applications provide scenarios that are catered to students' unique professional or office settings such as communicating technical project specifications to stakeholders, or negotiating deadlines with project managers. This presents a gap for students working in multinational and multilingual environments. Here, a student must not only be able to fluently communicate in the target language general scenarios such as greetings, but also more technical, domain-specific discussions about their work. The latter may not always be covered by the current general scenarios which these applications generate.

In this workshop contribution, we conduct an exploratory study and provide a preliminary evaluation on Duolingo's AI-generated lessons \cite{duolingo-ai} from the perspective of its users on how it affects their learning experience. With this understanding, we provide design considerations for language acquisition technologies that better cater to a student's individual professional circumstances. Specifically, we ask the following research questions (RQs): 

\textbf{RQ1}: What are the differences in perceptions of language learning students towards general and work-specific scenarios from Duolingo's AI-generated lessons?, and 

\textbf{RQ2}: How do these perceptions affect their learning experience towards gaining professional level fluency in their target language?

We conducted a formative study that surveyed five language learners employed by a multinational corporation about their experiences with Duolingo. They were asked questions about how often they encounter lessons containing scenarios that they directly experience in their work-related and general, non-work-related communications, and how these lessons affect their learning experience. They also provided specific lesson scenarios that would help them better gain professional-level communication fluency in their target language. We defined professional fluency as a level where \textit{an individual can comfortably communicate various work-related topics in their target language.}

Our findings reveal that general scenarios provide a valuable learning experience for language learners, particularly for beginners. Their simplicity and relatability allow them to easily learn the foundational aspects of their target language, such as grammar and culture. However, work-related scenarios, especially ones that contain domain specific information, provide an opportunity for language learners to bridge the gap between the current fluency and professional-level fluency. Lastly, language learners expressed their desire for more personalization that caters to their learning goals and experience.

\section{Background}


\subsection{Intelligent Tutoring Systems for Personalized Learning}

Intelligent tutoring systems (ITSs) are computerized teaching tools that mimic human teaching behavior by using techniques from AI, cognitive science, and educational research. \cite{nwana1990intelligent}. These ITSs are able to offer more personalized teaching methods because of the following modules that comprise its architecture: The \textit{expert knowledge module} is the part of the system that contains domain-specific information and is responsible for generating lessons that are to be taught to the student \cite{nwana1990intelligent, polson2013foundations}. The \textit{student model module} is responsible for diagnosing \cite{self1988studentmodels} the student's current understanding of the lesson material, and making necessary changes to the teaching medium to accommodate the student's needs \cite{polson2013foundations}. Early work by Self \cite{self1988studentmodels} formalized these as \textit{diagnostic} and \textit{strategic} functions, respectively. Lastly, the \textit{tutoring module} regulates the pedagogic interventions that will be presented to the student such as hints, tests, and explanations \cite{nwana1990intelligent, polson2013foundations}. ITSs can be very beneficial for a student. Prior work by \citet{its-meta} revealed that students who received tutoring from these ITSs outperformed students who didn't by a significant margin across different cultural and educational settings \cite{its-meta}. This is especially true more for local tests administered by specific instructional programs than standardized tests \cite{its-meta, reciprocal-teaching-meta, learning-programs-meta}.

\subsubsection{Duolingo as an Intelligent Tutoring System}

An article by \citet{duolingo-ai} describes Duolingo's process of creating courses and lessons. A set of human experts writes a prompt with a set of corresponding rules that would be provided to the LLM to generate an exercise in the target language. These lessons are then evaluated by a human "learning designer" to see if they align with the language and the culture they represent. In the context of an ITS, this would be the expert knowledge module.

\begin{figure}[h]
  \centering
  \includegraphics[width=\linewidth]{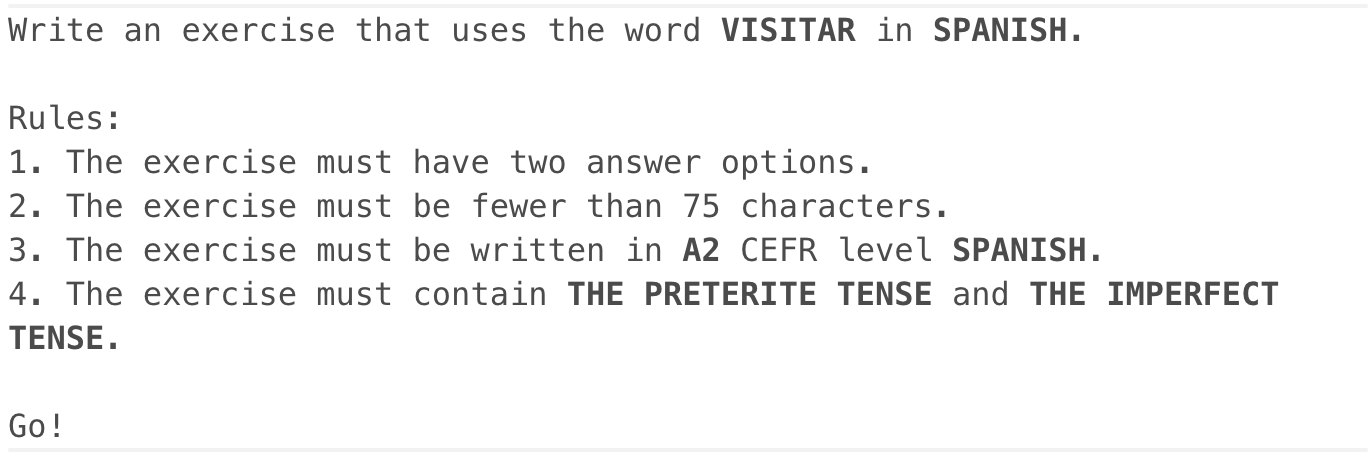}
  \caption{A sample process of how Duolingo's "learning designers" prompts an LLM to generate exercises for its courses. (Image was grabbed from https://blog.duolingo.com/large-language-model-duolingo-lessons/)}
  \Description{A sample process of how Duolingo's "learning designers" prompts an LLM to generate exercises for its courses}
\end{figure}

Duolingo also offers some personalization for its users. It contains a student model \cite{self1988studentmodels} called "Birdbrain" that infers a student's expertise and adjusts the lesson's difficulty level accordingly \cite{duolingo-birdbrain}, as well as provides certain lessons at specific phases of the learner's journey to maintain previously learned knowledge \cite{settles2016trainable}.

Lastly, Duolingo provides pedagogic interventions in the form of providing detailed feedback to the user when they answer lessons incorrectly \cite{duolingo-explanation}. This would be the tutoring module in an ITS.

\subsection{Second Language Acquisition Theory}

\textit{Second language acquisition} is a field of study that aims to understand how individuals acquire a second language. A prominent theory in the field is one by \citet{krashenpnp}. His theory comprises of five hypotheses, but emphasizes the input hypothesis as the most important concept as it attempts to answer how people acquire language. It claims that people acquire a language when the input is \textit{mostly} understood, meaning that there is some input that is beyond the current fluency level of the acquirer. The meaning of the input is not lost and is understood by the acquirer through context \cite{krashenpnp}.In the case of Duolingo, the lessons it generates serves as the input, and it provides the context through the real-world scenarios that it presents. It also adjusts the difficulty for each lesson such that new lessons are slightly beyond the user's current fluency level \cite{duolingo-difficulty}. 

In the realm of sociolinguistic theory, Dell Hymes developed the \textit{Communicative Competence Theory}. \citet{hymes1972communicative} posited that while grammatical knowledge of a language is important for an individual's acquisition, how an individual uses grammatical tools to construct sentences and participate in discourse is just as crucial. \citet{hymes1972communicative} accounted for the differences in language use that occur because of both the context it is used in, and the varying background of its language users \cite{whyte2019revisiting}. It is then, through the Communicative Competence Theory, that the measurement of a language learner's proficiency is not only based on their grammatical knowledge and syntactically accurate sentences, but also on how well they were able to communicate in different scenarios. 

The work of \citet{hymes1972communicative} led to a shift in the way second language acquisition was conducted. Initially, second language teachers emphasized grammatical structure and linguistic prescriptivism in their classrooms. This meant that the teacher's main goal was to pass on knowledge of the "right" use of the language. Following the propagation of Communicative Competence Theory, language teaching and learning eventually included examining what words, phrases, and structures were relevant to the context in which learners were speaking \cite{LILLIS2006666}. Social and hierarchical relationships, professional fields, and text types/genres (literary, academic, etc.) are just some examples of the different non-linguistic factors that can affect language use. By looking into how environment and context shape human interaction, the Communicative Competence Theory challenged what was deemed "appropriate" language use, by extending this definition to include both an utterances adherence to grammatical rules of a language and how well it communicated meaning based on social, professional, and even pragmatic context.

\section{Method}

We conducted a survey aiming to understand language learners' experiences with Duolingo. Duolingo was selected due to its popularity in the Philippines \cite{duolingo-philippines}. The survey was deployed on Qualtrics, and distributed through a multinational company based in the Philippines' communication channels. We received five valid responses. All participants were software engineers recruited from the company's Korean language class. The survey begins by asking participants about how long and how frequent they have used Duolingo for second language acquisition. Then for both general and work-related lesson scenarios provided by Duolingo, the survey asks how often they encounter them, and if they enhance or hamper their overall experience towards gaining professional-level fluency. Lastly, the participants suggested lesson scenarios that would enhance their learning experience.

\section{Findings and Discussion}

To better visualize our findings, we separate our design considerations according to our respondents' perceptions between lessons that contain \textit{general scenarios} and \textit{work-related scenarios}. We also describe some suggestions from our respondents on what type of lessons scenarios would help the easily gain professional-level fluency.

\subsection{General scenarios serve their purpose for setting the foundation for the language acquisition}

In the context of our study, general scenarios remain integral to participants' second language acquisition goals as all of them, especially self-described novices in the target language, report that these types of scenarios enhance their learning experience. Because these scenarios are relatable, and are encountered more frequently in their daily lives, it enabled them to grasp foundational language concepts such as grammar and vocabulary easily. This is shown in some responses to Q7: \textit{"The non-work-related lessons help in adding to my overall learning, specifically with grammar and vocabulary."} and \textit{"I can identify the words I hear during daily conversations"}. Duolingo currently satisfies this, as all respondents reported encountering these general scenarios more often than work-specific scenarios.

\subsection{Work-related scenarios serve as an opportunity to bridge the fluency gap between novice and professional}

We find that respondents encounter work-related scenarios much less frequently than general ones. One respondent even reported having never encountered any, but for those who did, the general perception is that the work-specific scenarios are able to bridge the fluency gap between novice and professional. One respondent reported that learning about work-specific scenarios helped him/her better understand culture and language. In a more specific instance, one respondent, who we presume is employed as a software developer, reported that he/she has yet to encounter scenarios that contain technical jargon such as CI/CD pipeline, user interface, and defect, widening his/her perceived learning gap towards being able to communicate in professional settings.  


However, we propose that these work-related scenarios may be more beneficial in later stages of their learning journey. One respondent noted that work-related scenarios are still not applicable at the beginner level, showing that students are aware that it is necessary to understand the fundamentals of the language from general scenarios before they can acquire professional-level fluency from work-related scenarios.

\subsection{Divergent topics on suggested lesson scenarios show a desire for a more personalized learning experience}

In our survey, we asked for suggestions on what type of lessons would help them gain professional level fluency. The suggestions' topics were divergent, such as: more language fundamentals, scenarios on negotiating task deadlines, traveling to the target language's country, and more everyday conversations. These topics suggest that there is an opportunity to leverage participants' backgrounds, goals, and habits to design language learning applications that can create more personalized lesson scenarios. Because Duolingo's Birdbrain/student modules only adjusts for the difficulty level, it does not adjust according to the lesson content that the individual learner desires.

\section{Limitations and Future Work}

Our work contains limitations. We recognize the small pool of participants, and plan to continue this study by recruiting more participants to strengthen our findings. For our future experiment, we plan to fine-tune an LLM to generate lessons that are more applicable in the technology industry. We will then conduct a long-term between-subjects study with software engineers as language learners. Our control group would be one without the ITS, the other group would use Duolingo, and the last one would use our fine-tuned LLMs. We will then compare language test scores and reported user experience between the groups.

\section{Conclusion}

We present a formative study evaluating Duolingo's LLM-generated lessons from the language learners' perspective. We found that learners perceive lessons with general scenarios as relatable and, therefore, helpful for them as novices to learn more foundational concepts of the language through immersion. However, the benefits of these general scenarios may not transfer well for their professional career settings, where scenarios typically involve very domain-specific technical conversations. It would be beneficial for language learning applications to provide both general scenarios and adapt to these types of individual work-specific scenarios to provide a more personalized learning experience in general. In closing, we envision an intelligent language tutoring system that is agentic. One that is able to understand and adapt to the user's changing background, goals, context, and environment, to create lesson content that caters to the each unique individual learner.

\section{Appendices}

\subsection{Survey Questions}

\begin{enumerate}
  \item[\textbf{Q1}] (Single Choice) How long have you been using Duolingo for second language acquisition?
  \begin{itemize}
     \item < 1 year
     \item 1-5 years
     \item 6-10 years
     \item > 10 years
   \end{itemize}
  \item[\textbf{Q2}] (Single Choice) How frequently do you use Duolingo for second language acquisition?
  \begin{itemize}
     \item Less than once a month
     \item Once a month
     \item 2-3 times a month
     \item Once a week
     \item 2-3 times a week
     \item Daily
   \end{itemize}
  \item[\textbf{Info:}] In language learning settings, real-world scenarios in the target language are commonly presented to the learners, providing a level of immersion to help them gain professional-level fluency in the target language.

  (we define professional-level fluency as where an individual who is able to comfortably communicate in the target language various work-related matters)
  \item[\textbf{Q3}] (Single Choice) How frequently do you encounter lessons containing scenarios that you directly experience in your work-related communications?
  \begin{itemize}
     \item Always
     \item Very Often
     \item About half the time
     \item Sometimes
     \item Never
   \end{itemize}
    \item[\textbf{Q4}] (Single Choice) Do these work-related lessons enhance/hamper your overall experience towards gaining professional-level fluency in your target language?
    \begin{itemize}
     \item Greatly enhances
     \item Somewhat enhances
     \item Neutral
     \item Somewhat hampers
     \item Greatly hampers
   \end{itemize}
   \item[\textbf{Q5}](Open Ended) Please provide a brief explanation on why you responded such in the previous question
   \item[\textbf{Q6}] (Single Choice) How frequently do you encounter lessons containing scenarios that you experience in your non-work-related communications?
    \begin{itemize}
     \item Always
     \item Very Often
     \item About half the time
     \item Sometimes
     \item Never
   \end{itemize}
   \item[\textbf{Q7}] (Single Choice) Do these non-work-related lessons enhance/hamper your overall experience towards gaining professional-level fluency in your target language?
    \begin{itemize}
     \item Greatly enhances
     \item Somewhat enhances
     \item Neutral
     \item Somewhat hampers
     \item Greatly hampers
   \end{itemize}
   \item[\textbf{Q8}](Open Ended) Please provide a brief explanation on why you responded such in the previous question
   \item[\textbf{Q9}] (Open Ended) What type of lesson scenario/s would help you easily gain professional level fluency in your target language?
\end{enumerate}

\section{Appendices}

\subsection{Survey Results}

\begin{figure}[h]
  \centering
  \includegraphics[width=\linewidth]{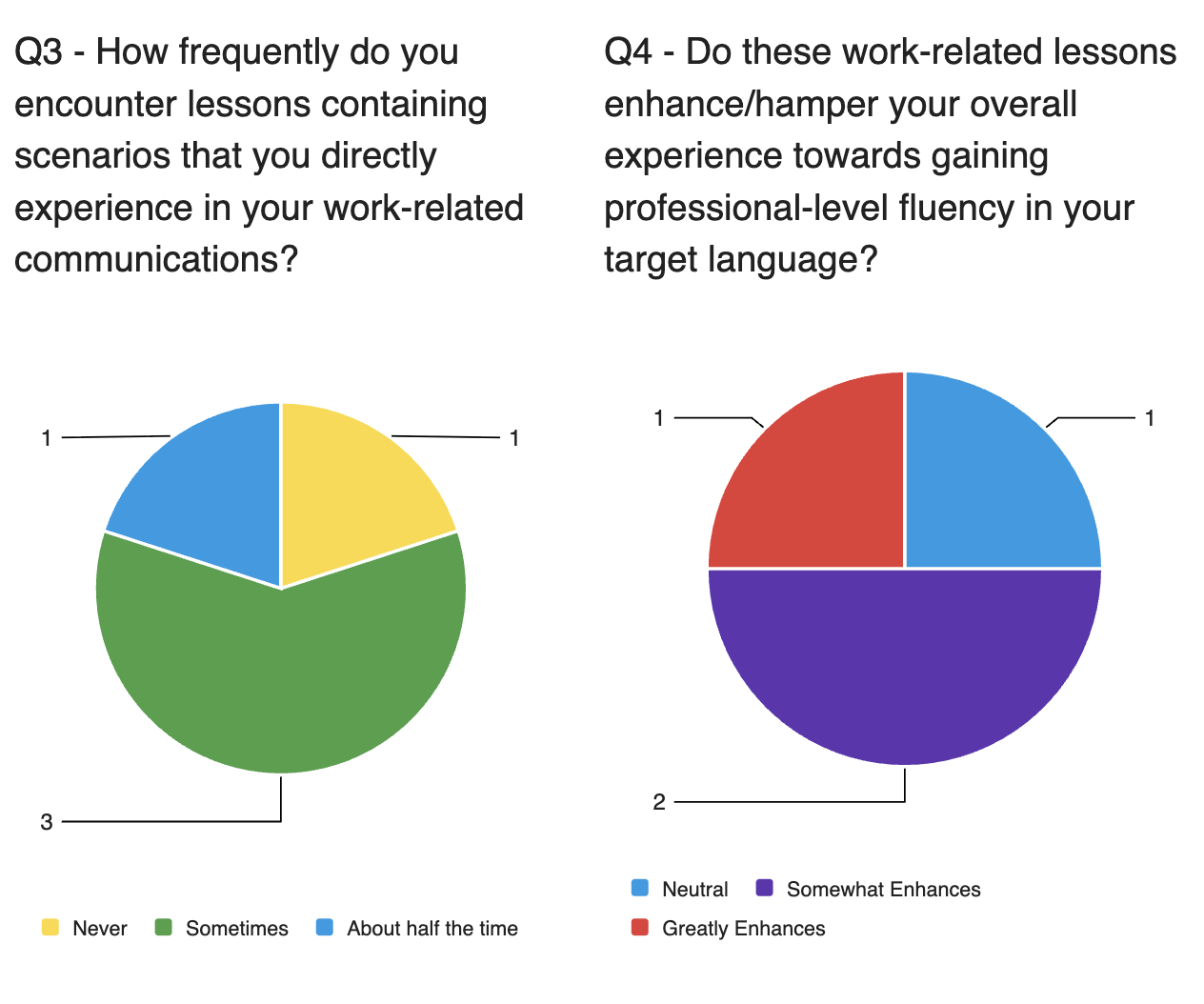}
  \caption{Survey results of questions on language learners' perceptions on lessons containing work-related scenarios)}
  \Description{Survey results of questions on language learners' perceptions on lessons containing work-related scenarios}
\end{figure}

\begin{figure}[h]
  \centering
  \includegraphics[width=\linewidth]{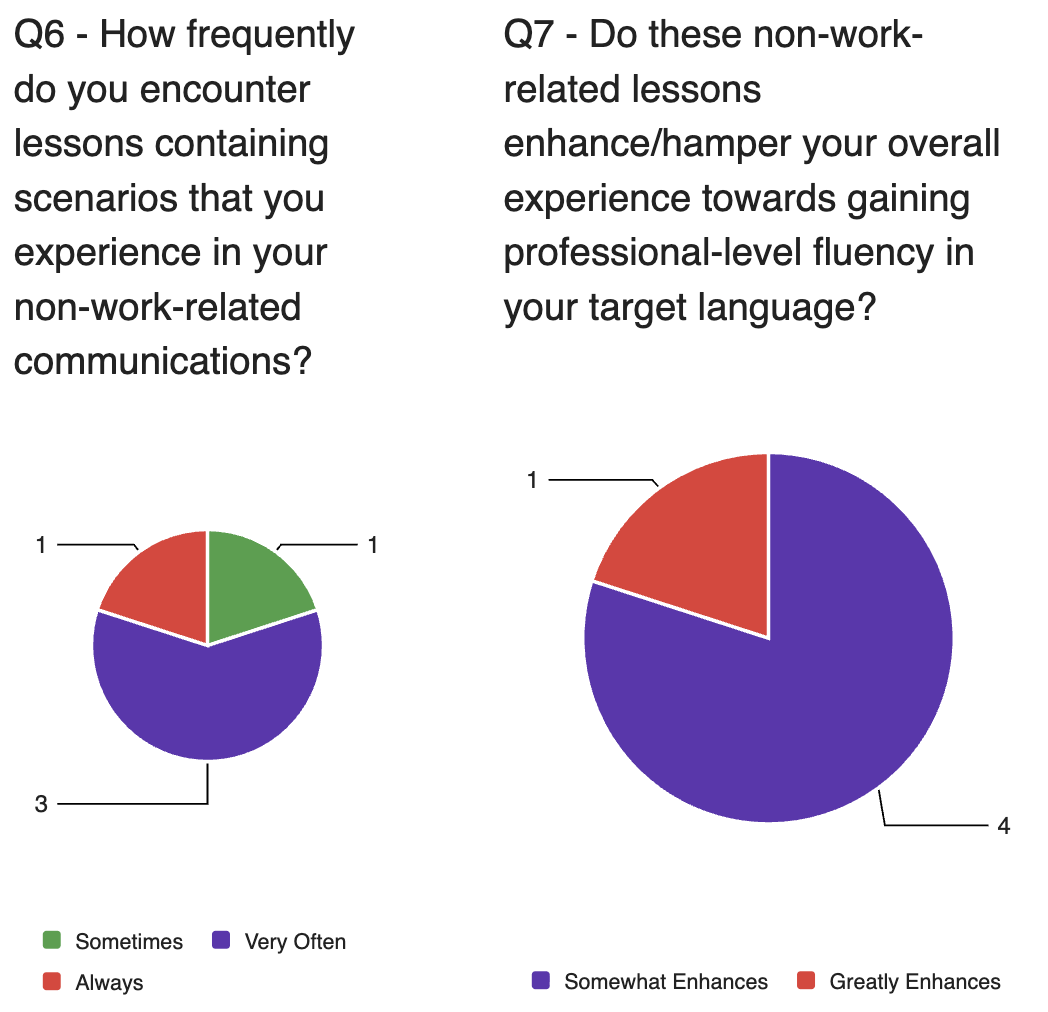}
  \caption{Survey results of questions on language learners' perceptions on lessons containing general scenarios)}
  \Description{Survey results of questions on language learners' perceptions on lessons containing general scenarios}
\end{figure}

\bibliographystyle{ACM-Reference-Format}
\bibliography{sample-base}










\end{document}